\documentclass[11pt,a4paper]{article}
\usepackage[hyperref]{emnlp2018}
\usepackage{times}
\usepackage{latexsym}
\usepackage{multirow}
\usepackage{times}
\usepackage{latexsym}
\usepackage{amsmath,bm}
\usepackage{amssymb}
\usepackage{dsfont}
\usepackage{booktabs}
\usepackage{graphicx}
\usepackage{colortbl}
\usepackage{url}
\usepackage{todonotes}
\usepackage{xspace}
\usepackage{dblfloatfix}
\usepackage{pifont}
\usepackage{arydshln}
\usepackage{csquotes}
\usepackage{svg}
\usepackage[nomessages]{fp}
\usepackage{siunitx}

\newcommand\mc[1]{\multicolumn{1}{c}{#1}} 
\newcommand\mcc[1]{\multicolumn{1}{c:}{#1}} 

\newcommand{\rowc}{black!10}

\newcommand{\Reisinger}{\textsc{lr}\xspace}
\newcommand{\AAAI}{\textsc{crf}\xspace}
\newcommand{\SprOneRand}{\textsc{spr1-rand}}
\newcommand{\SprOneB}{\textsc{spr1}}
\newcommand{\SprOneBSprTwo}{\textsc{\SprOneB+2}}
\newcommand{\mtSprOneB}{\textsc{mt:\SprOneB}}
\newcommand{\pbSprOneB}{\textsc{pb:\SprOneB}}
\newcommand{\mtpbSprOneB}{\textsc{mt:pb:\SprOneB}}
\newcommand{\SprOneBSupersense}{\textsc{\SprOneB+\supersense}}
\newcommand{\mtSprOneBSprTwo}{\textsc{mt:\SprOneBSprTwo}}
\newcommand{\mtSprOneBSupersense}{\textsc{mt:\SprOneBSupersense}}
\newcommand{\psopt}{\textsc{ps-ms}}

\newcommand{\SprOneS}{\textsc{spr1s}}
\newcommand{\mtSprOneS}{\textsc{mt:\SprOneS}}
\newcommand{\pbSprOneS}{\textsc{pb:\SprOneS}}
\newcommand{\mtpbSprOneS}{\textsc{mt:pb:\SprOneS}}
\newcommand{\SprTwo}{\textsc{spr2}}
\newcommand{\mtSprTwo}{\textsc{mt:\SprTwo}}
\newcommand{\pbSprTwo}{\textsc{pb:\SprTwo}}
\newcommand{\mtpbSprTwo}{\textsc{mt:pb:\SprTwo}}

\newcommand{\SprOne}{\textsc{spr1}}

\newcommand{\mtSprOne}{\textsc{mt:\SprOne}}
\newcommand{\pbSprOne}{\textsc{pb:\SprOne}}
\newcommand{\mtpbSprOne}{\textsc{mt:pb:\SprOne}}
\newcommand{\supersense}{\textsc{wsd}}
\newcommand{\SprOneSupersense}{\textsc{\SprOne+\supersense}}
\newcommand{\SprOneSprTwo}{\textsc{\SprOne+\SprTwo}}
\newcommand{\mtSprOneSupersense}{\textsc{mt:\SprOneSupersense}}
\newcommand{\mtSprOneSprTwo}{\textsc{mt:\SprOneSprTwo}}

\usepackage{url}

\aclfinalcopy 

\title{Neural-Davidsonian Semantic Proto-role Labeling}

\newcommand{\xmark}{\ding{55}}%
  
\author{Rachel Rudinger\\
  Johns Hopkins University \\\And
  Adam Teichert\\
  Johns Hopkins Univeristy\\\AND
  Ryan Culkin\\
  Johns Hopkins University\\\And
  Sheng Zhang\\
  Johns Hopkins University\\\And
  Benjamin Van Durme\\
  Johns Hopkins University
  }

\date{}

\begin{document}
\maketitle
\begin{abstract}
We present a model for semantic proto-role labeling (SPRL) using an adapted bidirectional LSTM encoding strategy that we call \emph{Neural-Davidsonian}: predicate-argument structure is represented as pairs of hidden states corresponding to predicate and argument head tokens of the input sequence. We demonstrate: (1) state-of-the-art results in SPRL, and (2) that our network naturally shares parameters between  attributes, allowing for learning new attribute types with limited added supervision.
\end{abstract}
\section{Introduction}
\label{sec:intro}
Universal Decompositional Semantics (UDS) \cite{uds2016} is a contemporary semantic representation of text  \cite{abend2017state} that forgoes traditional inventories of semantic categories in favor of bundles of simple, interpretable properties. In particular, UDS includes a practical implementation of Dowty's theory of \textit{thematic proto-roles} \cite{dowty_thematic_1991}: arguments are labeled with properties typical of Dowty's \textit{proto-agent} (\textsc{awareness}, \textsc{volition} ...) and \textit{proto-patient} (\textsc{changed state} ...).

Annotated corpora have allowed the exploration of \textit{Semantic Proto-role Labeling} (SPRL)
\footnote{SPRL and SPR refer to the labeling task and the underlying semantic representation, respectively.}
as a natural language processing task
\cite{TACL674,uds2016,teichert2017sprl}.
For example, consider the following sentence, in which a particular pair of predicate and argument heads have been emphasized: \enquote{The cat {\em ate} the \underline{rat}.}
An SPRL system must infer from the context of the sentence whether the
\underline{rat} had {\sc volition}, {\sc changed-state}, and {\sc
  existed-after} the {\em eating} event (see Table
\ref{tab:spr1_breakdown_test} for more properties).

\begin{figure}[t]
  \centering
 \fbox{\includegraphics[width=0.4\textwidth]{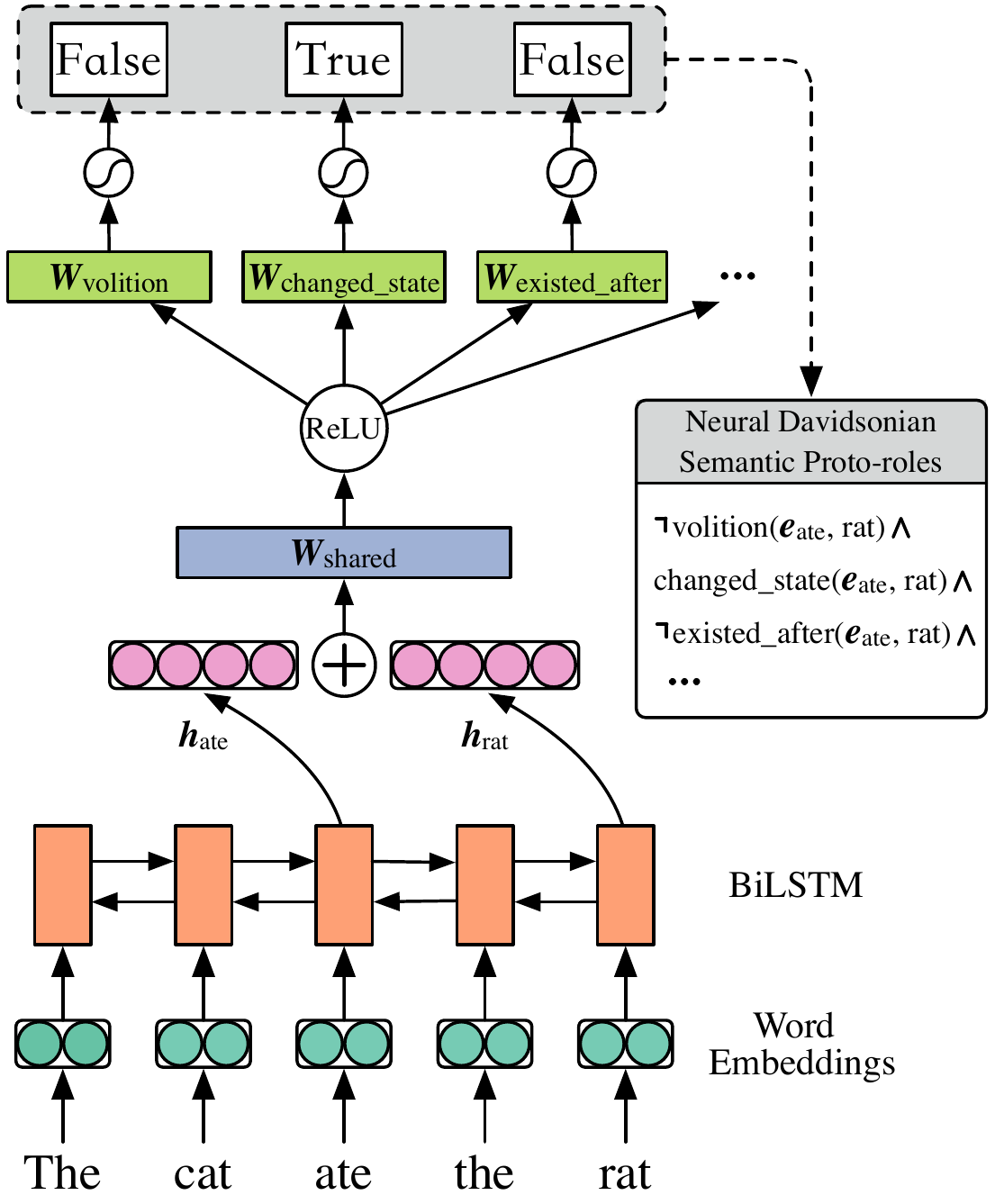}}
 \caption{BiLSTM sentence encoder with SPR decoder. Semantic proto-role labeling is with respect to a specific predicate and argument within a sentence, so the decoder receives the two corresponding hidden states.}
  \label{fig:spr_net}
\end{figure}
\vspace{-1mm}

We present an intuitive neural model that achieves
state-of-the-art performance for SPRL.\footnote{Implementation available at \url{https://github.com/decomp-sem/neural-sprl}.}  As depicted in
Figure \ref{fig:spr_net}, our model's architecture is an extension of
the bidirectional LSTM, capturing a Neo-Davidsonian 
like intuition, wherein select pairs
of hidden states are concatenated to yield a dense representation of
predicate-argument structure and fed to a prediction layer for end-to-end
training. We include a thorough quantitative analysis highlighting
the contrasting errors between the proposed model and previous
(non-neural) state-of-the-art.

\begin{table*}[]
\centering
\begin{tabular}{llcc}
\toprule
SPR Property & Explanation of Property &  \includegraphics[scale=0.01]{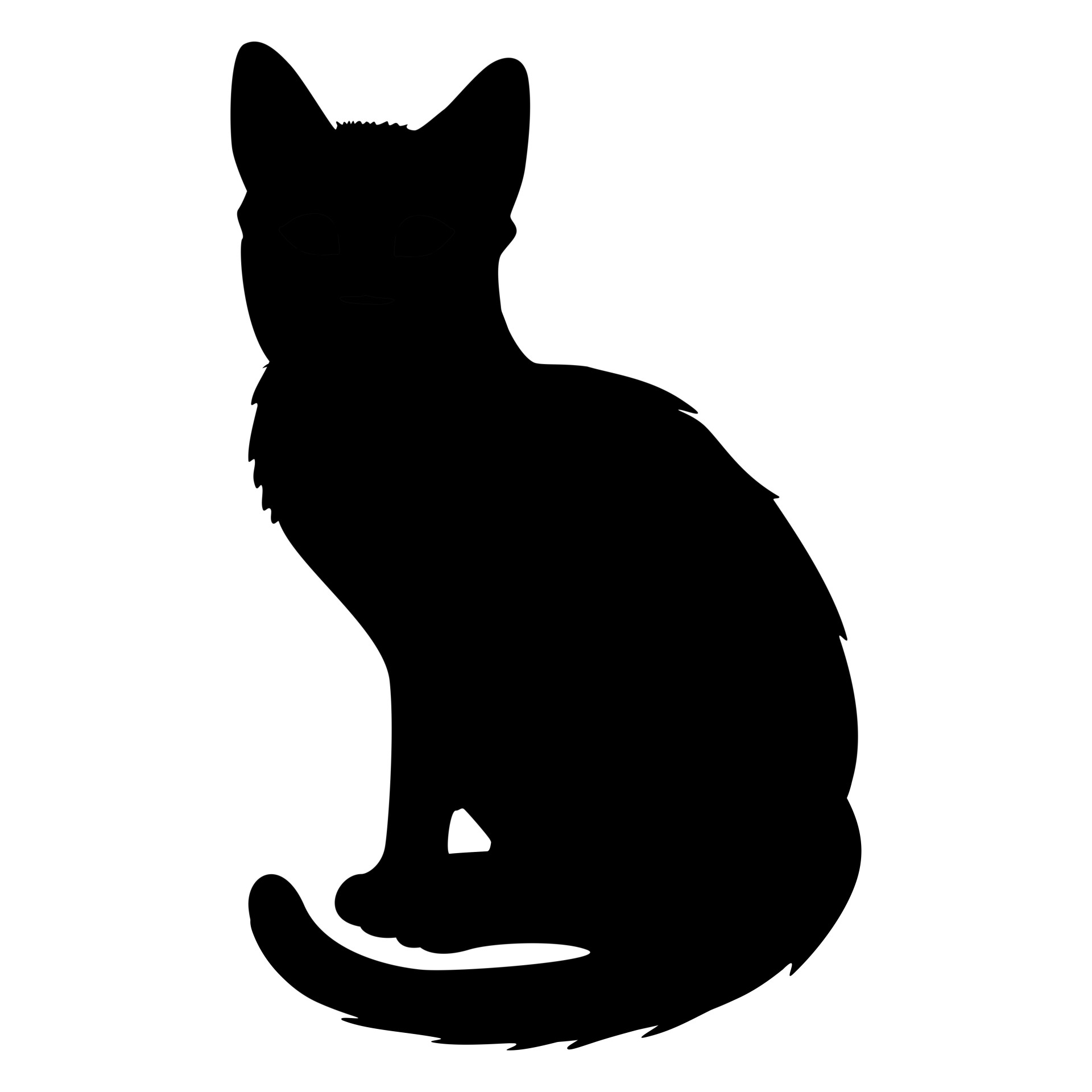}   &  \includegraphics[scale=0.011]{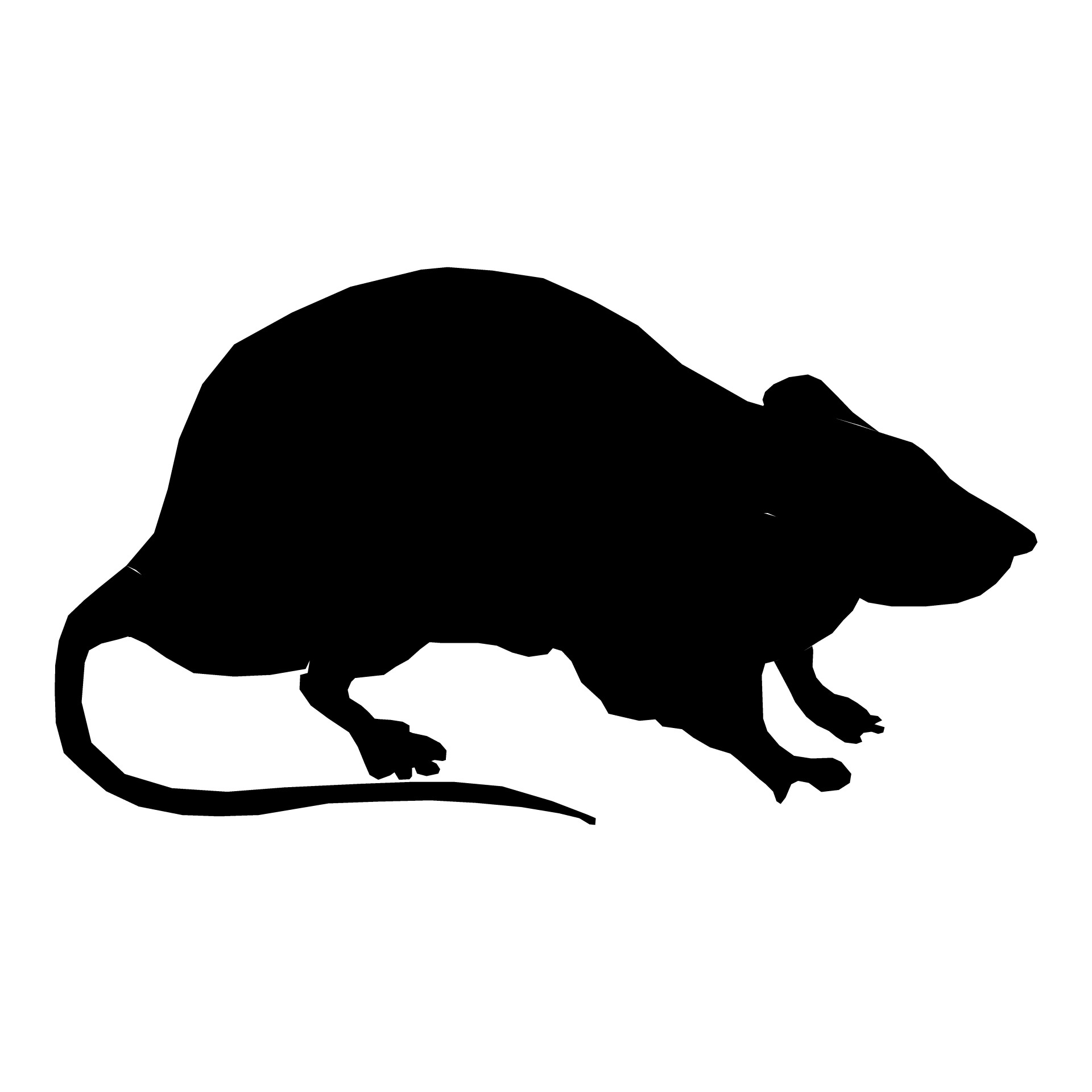}   \\ \midrule
\textsc{instigation}        & \texttt{Arg} caused the \texttt{Pred} to happen?                                                                                          & \checkmark  & \xmark \\ 
\textsc{volitional}         & \texttt{Arg} chose to be involved in the \texttt{Pred}?                                                                                   & \checkmark  & \xmark \\ 
\textsc{aware}              & \begin{tabular}[c]{@{}l@{}}\texttt{Arg} was/were aware of being\\ involved in the \texttt{Pred}?\end{tabular}                             & \checkmark  & \checkmark  \\ 
\textsc{physically existed} & \texttt{Arg} existed as a physical object?                                                                                       & \checkmark  & \checkmark  \\ 
\textsc{existed after}      & \texttt{Arg} existed after the \texttt{Pred} stopped?                                                                                     & \checkmark  & \xmark \\ 
\textsc{changed state}      & \begin{tabular}[c]{@{}l@{}}The \texttt{Arg} was/were altered or somehow\\ changed during or by the end of the \texttt{Pred}?\end{tabular} & \checkmark & \checkmark  \\ \bottomrule
\end{tabular}
\caption{\label{tab:sprexample} Example SPR annotations for the toy example ``The cat ate the rat,'' where the \texttt{Predicate} in question is ``\textit{ate}'' and the \texttt{Argument} in question is either ``cat'' or ``rat.'' Note that not all SPR properties are listed, and the binary labels (\checkmark, \xmark) are coarsened from a 5-point Likert scale.}
\vspace{-.5em}
\end{table*}

In addition, our network naturally shares a subset of parameters between attributes. We demonstrate how this allows learning to predict new attributes with limited supervision: a key finding that could support efficient expansion of new SPR attribute types in the future.

\section{Background}
\label{sec:background}

\newcite{davidson-67} is credited for representations of meaning  involving propositions composed of a fixed arity predicate, all of its core arguments arising from the natural language syntax, and a distinguished event variable.  The earlier example could thus be
denoted (modulo tense) as $(\exists e)\textbf{eat}[(e,\textsc{cat}, \textsc{rat})]$, where the variable $e$ is a \textit{reification} of the eating event.
The order of the arguments in the predication implies their role,
where leaving arguments unspecified (as in ``The cat eats'') can be
handled either by introducing variables for unstated arguments, e.g., $(\exists e)(\exists x)[\textbf{eat}(e,\textsc{cat}, x)]$, or by creating new predicates that correspond to different arities, e.g., $(\exists e)\textbf{eat\_intransitive}[(e,\textsc{cat})]$.\footnote{This
formalism aligns with that used in PropBank
\cite{palmer2005_propbank}, which associated numbered, core arguments with each sense of a verb in their corpus annotation.}    The Neo-Davidsonian
approach \cite{castaneda-67,parsons1995thematic}, which we follow in this work, allows for variable arity by mapping the argument positions of individual predicates to generalized \emph{semantic roles}, shared across predicates,\footnote{For example, as seen in FrameNet~\cite{baker1998berkeley}.} e.g., {\sc agent}, {\sc patient} and {\sc theme}, in: $(\exists e)[\textbf{eat}(e) \wedge
\textbf{Agent}(e,\textsc{cat}) \wedge \textbf{Patient}(e,\textsc{rat})]$.

\newcite{dowty_thematic_1991} conjectured that the distinction between
the role of a prototypical {\bf Agent} and prototypical {\bf Patient}
could be decomposed into a number of semantic properties such as \emph{``Did
the argument change state?''}. Here we formulate this as a Neo-Davidsonian representation employing \emph{semantic proto-role} (SPR) attributes:
\begin{align*}
        (\exists e)\left[\right.&\textbf{eat}(e) & \\
& \wedge \textbf{volition}(e,\textsc{cat}) \wedge \textbf{instigation}(e,\textsc{cat}) ...\\
& \wedge \neg\textbf{volition}(e,\textsc{rat}) \wedge \textbf{destroyed}(e,\textsc{rat})...\left.\right]
\end{align*}

Dowty's theory was empirically verified by \newcite{kako2006thematic}, followed by pilot~\cite{madnani-2010} and large-scale~\cite{TACL674} corpus annotation efforts, the latter introducing a logistic regression baseline for SPRL. \newcite{teichert2017sprl} refined the evaluation protocol,\footnote{Splitting train/dev/test along Penn Treebank boundaries and casting the SPRL task as multi-label binary classification.} and developed a CRF~\cite{lafferty2001conditional} for the task, representing existing state-of-the-art.

Full details about the SPR datasets introduced by \newcite{TACL674} and \newcite{uds2016}, which we use in this work, are provided in Appendix \ref{sec:data}. 
For clarity, Table \ref{tab:sprexample} shows a toy SPRL example, including a few sample SPR properties and explanations.

\section{``Neural-Davidsonian'' Model}
\label{sec:model}
Our proposed SPRL model (Fig.~\ref{fig:spr_net})  determines the value of each attribute (e.g., {\sc volition}) on
an \textit{argument} ($a$) with respect to a particular \textit{predication} ($e$) as a function on the latent states associated with the pair, $(e,a)$, in the context of
a full sentence.  Our architecture encodes the sentence using a
shared, one-layer, bidirectional LSTM
\cite{hochreiter_long_1997,graves_hybrid_2013}.  We then obtain a
continuous, vector representation
$\bm{h}_{ea}=\left[\bm{h}_e;\bm{h}_a\right]$, for each
predicate-argument pair as the concatenation of the hidden BiLSTM
states $\bm{h}_e$ and $\bm{h}_a$ corresponding to the syntactic head
of the predicate of \textit{e} and argument \textit{a} respectively. These
heads are obtained over gold syntactic parses using
the predicate-argument detection tool, PredPatt \cite{uds2016}.\footnote{Observed to be state-of-the-art by~\newcite{zhang-EtAl:2017:IWCS}.}

For each SPR attribute, a score
is predicted by passing $\bm{h}_{ea}$ through a separate
two-layer perceptron, with the weights of the first layer  shared across
all attributes:
\begin{align*}
  \text{Score}(\text{attr},\bm{h}_{ea})=\bm{W}_{\text{attr}}\left[g\left(\bm{W}_{\text{shared}}\left[\bm{h}_{ea}\right]\right)\right]
\end{align*}

This architecture accomodates the definition of SPRL as multi-label
binary classification given by \newcite{teichert2017sprl} by treating the
score as the log-odds of the attribute being present (i.e.
$\text{P}(\text{attr}|\bm{h}_{ea})=\frac{1}{1+\exp[-
\text{Score}(\text{attr},\bm{h}_{ea})]}$). This architecture also supports SPRL as a {\em scalar} regression task where the parameters of the network are tuned to directly minimize the discrepancy between the predicted score and a reference scalar label.
The loss for the binary and scalar models are negative log-probability and squared error, respectively; the losses are summed over all SPR attributes.\\

\noindent{\bf Training with Auxiliary Tasks}
A benefit of the shared neural-Davidsonian representation is that it
offers many levels at which multi-task learning may be leveraged to
improve parameter estimation so as to produce semantically rich
representations $\bm{h}_{ea}$, $\bm{h}_{e}$, and $\bm{h}_{a}$.  For
example, the sentence encoder might be pre-trained as an encoder for
machine translation, the argument representation $\bm{h}_{a}$ can be
jointly trained to predict word-sense, the predicate
representation, $\bm{h}_{e}$, could be jointly trained to predict
factuality \cite{Sauri2009,rudinger-EtAl:2018:N18}, and the predicate-argument representation, $\bm{h}_{ea}$,
could be jointly trained to predict other semantic role formalisms (e.g. PropBank SRL---suggesting a neural-Davidsonian {\em SRL} model in contrast to recent BIO-style neural models of SRL \cite{he-EtAl:2017:Long3}).

To evaluate this idea empirically, we experimented with a number of multi-task
training strategies for SPRL.  
While all settings outperformed prior work in
aggregate, simply initializing the BiLSTM parameters with a pretrained English-to-French machine
translation encoder\footnote{using a modified version of \texttt{OpenNMT-py} \cite{opennmt} trained on the
  $10^9$ Fr-En corpus
  \cite{callisonburch-EtAl:2009:WMT-09} (Appendix A).} produced the best
results,\footnote{e.g. this initialization resulted in raising
  micro-averaged F1 from 82.2 to 83.3} so we simplify discussion by focusing on that model.
The efficacy of MT pretraining that we observe here comes as no surprise given prior work demonstrating, e.g., the utility of bitext for paraphrase \cite{ganitkevitch-vandurme-callisonburch:2013:NAACL-HLT}, that NMT pretraining yields improved contextualized word embeddings\footnote{More recent discoveries on the usefulness of language model pretraining \cite{peters-EtAl:2018:N18-1,howard-ruder:2018:Long} for RNN encoders suggest a promising direction for future SPRL experiments.} \cite{NIPS2017_7209}, and that NMT encoders specifically capture useful features for SPRL \cite{poliak2018evaluation}.

Full details about each multi-task experiment, including a full set of ablation results, are reported in Appendix \ref{sec:multi}; details about the corresponding datasets are in Appendix \ref{sec:data}.

Except in the ablation experiment of Figure \ref{fig:ablation}, our model was trained on
only the SPRL data and splits used by \newcite{teichert2017sprl}
(learning all properties jointly), using
GloVe\footnote{300-dimensional, uncased; \texttt{glove.42B.300d} from
  \url{https://nlp.stanford.edu/projects/glove/}; 15,533
  out-of-vocabulary words across all datasets were assigned a random
  embedding (uniformly from $\left[-.01,.01\right]$). Embeddings
  remained fixed during training.} embeddings and with the
MT-initialized BiLSTM. Models were implemented in PyTorch and trained
end-to-end with Adam optimization \cite{DBLP:journals/corr/KingmaB14}
and a default learning rate of $10^{-3}$. Each model was trained for
ten epochs, selecting the best-performing epoch on dev.\\

\noindent{\bf Prior Work in SPRL}
We additionally include results from prior work: ``\Reisinger'' is the
logistic-regression model introduced by \newcite{TACL674} and
``\AAAI'' is the CRF model (specifically \texttt{SPRL}$^\star$) from
\newcite{teichert2017sprl}.
Although \newcite{uds2016} released additional SPR annotations, we are unaware
of any benchmark results on that data; however, our multi-task results in
Appendix \ref{sec:multi} do use the data and we find
(unsurprisingly) that concurrent training on the two SPR datasets can
be helpful. Using only data and splits from \newcite{uds2016}, the
scalar regression architecture of Table \ref{tab:spr1c_breakdown_test}
achieves a Pearson's $\rho$ of 0.577 on test.

There are a few noteworthy differences between  our neural model and the CRF of prior work.
As an adapted BiLSTM, our model easily exploits the benefits of large-scale pretraining, in the form of GloVe embeddings and MT pretraining, both absent in the CRF.
Ablation experiments (Appendix \ref{sec:multi}) show the advantages conferred by these features.
In contrast, the discrete-featured CRF model makes use of gold dependency labels, as well as joint modeling of SPR attribute pairs with explicit joint factors, both absent in our neural model.
Future SPRL work could explore the use of models like the LSTM-CRF \cite{lample-EtAl:2016:N16-1,ma-hovy:2016:P16-1} to combine the advantages of both paradigms.

\section{Experiments}
\label{sec:experiments}
\begin{table}[t]
\begin{center}
\small
\begin{tabular}{lll:c|c}
\toprule
{} &  \multicolumn{2}{c}{previous work} & \multicolumn{2}{c}{this work} \\
{} &  \!\Reisinger\! & \!\AAAI\! & \!binary\! & \!scalar\! \\
\midrule
instigation                       &   76.7&         85.6&  \bf{88.6}&0.858 \\
\rowcolor{\rowc}volition          &   69.8&         86.4&  \bf{88.1}&0.882 \\
awareness                         &   68.8&         87.3&  \bf{89.9}&0.897 \\
\rowcolor{\rowc}sentient          &   42.0&         85.6&  \bf{90.6}&0.925 \\
physically existed                &   50.0&         76.4&  \bf{82.7}&0.834 \\
\rowcolor{\rowc}existed before    &   79.5&         84.8&  \bf{85.1}&0.710 \\
existed during                    &   93.1&         \bf{95.1}& 95.0 &0.673 \\
\rowcolor{\rowc}existed after     &   82.3&         \bf{87.5}& 85.9 &0.619 \\
created                           &   0.0&          \bf{44.4}& 39.7 &0.549 \\
\rowcolor{\rowc}destroyed         &   17.1&         0.0&   \bf{24.2}&0.346 \\
changed                           &   54.0&         67.8&  \bf{70.7}&0.592 \\
\rowcolor{\rowc}changed state     &   54.6&         66.1&  \bf{71.0}&0.604 \\
changed possession                &   0.0&          38.8&  \bf{58.0}&0.640 \\
\rowcolor{\rowc}changed location  &   6.6&          35.6&  \bf{45.7}&0.702 \\
stationary                        &   13.3&         21.4&  \bf{47.4}&0.711 \\
\rowcolor{\rowc}location          &   0.0&          18.5&  \bf{53.8}&0.619 \\
physical contact                  &   21.5&         40.7&  \bf{47.2}&0.741 \\
\rowcolor{\rowc}manipulated       &   72.1&         86.0&  \bf{86.8}&0.737 \\
\hline                                                              
micro f1                          &   71.0  &      81.7 &  \bf{83.3}& \\
\rowcolor{\rowc}macro f1        &55.4$^\star$&65.9$^\star$&\bf{71.1} &\\
macro-avg pearson                 &         &           &        & 0.753 \\
\bottomrule           
\end{tabular}
\end{center}
\caption{\label{tab:spr1_breakdown_test} SPR comparison to
  \newcite{teichert2017sprl}. Bold number indicate best F1 results in
  each row. Right-most column is pearson correlation coeficient for a
  model trained and tested on the scalar regression formulation of the
  same data.}
\vspace{-1em}
\end{table}

\begingroup
\renewcommand*{\arraystretch}{1.1}
\begin{table}[t]
\small
\centering
\begin{tabular}{@{}llrrr:rrr@{}} \toprule
&  &    \multicolumn{3}{c:}{phys. contact} & \multicolumn{3}{c}{volition} \\
  \midrule

  &  & \mc{\parbox[t]{1px}{\rotatebox[origin=l]{90}{\small \#~{\sc Differ}}}}
   & \mc{\parbox[t]{1px}{\rotatebox[origin=l]{90}{\small $\Delta$~{\sc False--}}}}
   & \mcc{\parbox[t]{1px}{\rotatebox[origin=l]{90}{\small $\Delta$~{\sc False+}}}}
   & \mc{\parbox[t]{1px}{\rotatebox[origin=l]{90}{\small \#~{\sc Differ}}}}
   & \mc{\parbox[t]{1px}{\rotatebox[origin=l]{90}{\small $\Delta$~{\sc False--}}}}
   & \mc{\parbox[t]{1px}{\rotatebox[origin=l]{90}{\small $\Delta$~{\sc False+}}}} \\
1  &   {\sc All} &\FPeval{\r}{round(27+13+17+23,0)}$\r$&\FPeval{\r}{round(-(27-13),0)}$\r$   & \FPeval{\r}{round(-(17-23),0)}$\r$&\FPeval{\r}{round(27+25+15+13,0)}$\r$& \FPeval{\r}{round(-(27-13),0)}$\r$ & \FPeval{\r}{round(-(25-15),0)}$\r$ \\
2  &   {\sc ProperNoun}   & \FPeval{\r}{round(6+4+5+3,0)}$\r$ & \FPeval{\r}{round(-(6-4  ),0)}$\r$   & \FPeval{\r}{round(-(5-3  ),0)}$\r$ & \FPeval{\r}{round(3+8+3+7,0)}$\r$  & \FPeval{\r}{round(-(3-7  ),0)}$\r$ & \FPeval{\r}{round(-(8-3  ),0)}$\r$ \\
3  &   {\sc Org.} & \FPeval{\r}{round(9+0+2+4,0)}$\r$ & \FPeval{\r}{round(-(9-0  ),0)}$\r$   & \FPeval{\r}{round(-(2-4  ),0)}$\r$ & \FPeval{\r}{round(13+6+5+7,0)}$\r$ & \FPeval{\r}{round(-(13-7 ),0)}$\r$ & \FPeval{\r}{round(-(6-5  ),0)}$\r$ \\
4  &   {\sc Pronoun}      & \FPeval{\r}{round(1+1+0+8,0)}$\r$ & \FPeval{\r}{round(-(1-1  ),0)}$\r$   & \FPeval{\r}{round(-(0-8  ),0)}$\r$ & \FPeval{\r}{round(3+3+3+3,0)}$\r$  & \FPeval{\r}{round(-(3-3  ),0)}$\r$ & \FPeval{\r}{round(-(3-3  ),0)}$\r$ \\
5  &   {\sc PhraseVerb}   & \FPeval{\r}{round(8+2+2+2,0)}$\r$ & \FPeval{\r}{round(-(8-2  ),0)}$\r$   & \FPeval{\r}{round(-(2-2  ),0)}$\r$ & \FPeval{\r}{round(4+2+3+0,0)}$\r$  & \FPeval{\r}{round(-(4-0  ),0)}$\r$ & \FPeval{\r}{round(-(2-3  ),0)}$\r$ \\
6  &   {\sc Metaphor}     & \FPeval{\r}{round(5+0+4+2,0)}$\r$ & \FPeval{\r}{round(-(5-0  ),0)}$\r$   & \FPeval{\r}{round(-(4-2  ),0)}$\r$ & \FPeval{\r}{round(2+2+2+0,0)}$\r$  & \FPeval{\r}{round(-(2-0  ),0)}$\r$ & \FPeval{\r}{round(-(2-2  ),0)}$\r$ \\
7  &   {\sc LightVerb}    & \FPeval{\r}{round(2+0+1+2,0)}$\r$ & \FPeval{\r}{round(-(2-0  ),0)}$\r$   & \FPeval{\r}{round(-(1-2  ),0)}$\r$ & \FPeval{\r}{round(2+0+2+1,0)}$\r$  & \FPeval{\r}{round(-(2-1  ),0)}$\r$ & \FPeval{\r}{round(-(0-2  ),0)}$\r$ \\
  \bottomrule
\end{tabular}

\caption{\label{tab:makes_phys_contact} Manual error analysis on a sample of instances (80 for each property) where outputs of \AAAI and the binary model from Table \ref{tab:spr1_breakdown_test} differ. {\bf Negative} $\Delta$ {\sc False+} and $\Delta$ {\sc False--} indicate the neural model represents a {\bf net {\em reduction} in type I and type II errors} respectively over \AAAI. Positive values indicate a net increase in errors. Each row corresponds to one of several (overlapping) subsets of the 80 instances in disagreement:
  (1) all (sampled) instances;
  (2) argument is a proper noun;
  (3) argument is an organization or institution;
  (4) argument is a pronoun;
  (5) predicate is phrasal or a particle verb construction;
  (6) predicate is used metaphorically;
  (7) predicate is a light-verb construction.  \#{\sc Differ} is the size of the respective subset.
}
\vspace{-1.2em}
\end{table}
\endgroup

Table \ref{tab:spr1_breakdown_test} shows a side-by-side comparison
of our model with prior work. The full breakdown of F1 scores over
each individual property is provided. For every property except
\textsc{existed during}, \textsc{existed after}, and
\textsc{created} we are able to exceed prior performance. For some
properties, the absolute F1 gains are quite large: \textsc{destroyed} (+24.2), \textsc{changed
possession} (+19.2.0), \textsc{changed location} (+10.1),
\textsc{stationary} (+26.0) and \textsc{location} (+35.3).
We also report performance with a scalar regression version of the model,
evaluated with Pearson correlation.
The scalar model is with respect to the original SPR annotations on a 5-point Likert scale, instead of a binary cut-point along that scale ($> 3$).

 \noindent\textbf{Manual Analysis } We select two properties
(\textsc{volition} and \textsc{makes physical contact}) to perform a
manual error analysis with respect to \AAAI~\footnote{We
obtained the \AAAI dev system predictions of
\newcite{teichert2017sprl} via personal communication with the
authors.} and our binary model from Table
\ref{tab:spr1_breakdown_test}.  For each property, we sample 40 dev
instances with gold labels of ``True'' ($>3$) and 40 instances of
``False'' ($\leq 3$), restricted to cases where the two system
predictions disagree.\footnote{According to the reference, of the
1071 dev examples, 150 have physical contact and 350 have
volition. The two models compared here differed in phy. contact on 62
positive and 44 negative instances and for volition on 43 positive and
54 negative instances.}  We manually label each of these instances
for the six features shown in Table \ref{tab:makes_phys_contact}.
For example, given the input \enquote{\underline{He} {\em sits} down at the piano and plays,} our neural model correctly predicts that \underline{He} makes physical contact during the {\em sitting}, while \AAAI does not.  Since \underline{He} is a pronoun, and {\em sits down} is phrasal, this example contributes $-1$ to $\Delta$~{\sc False--} in rows 1, 4 and 5.

For both properties our model appears more likely to correctly
classify the argument in cases where the predicate is a phrasal
verb. This is likely a result of the fact that the BiLSTM has stronger
language-modeling capabilities than the \AAAI, particularly with MT
pretraining. In general, our model increases the false-positive rate
for \textsc{makes physical contact}, but especially when the argument
is pronominal.\\

\begin{figure}[t]
\centering
  \includegraphics[width=.472\textwidth]{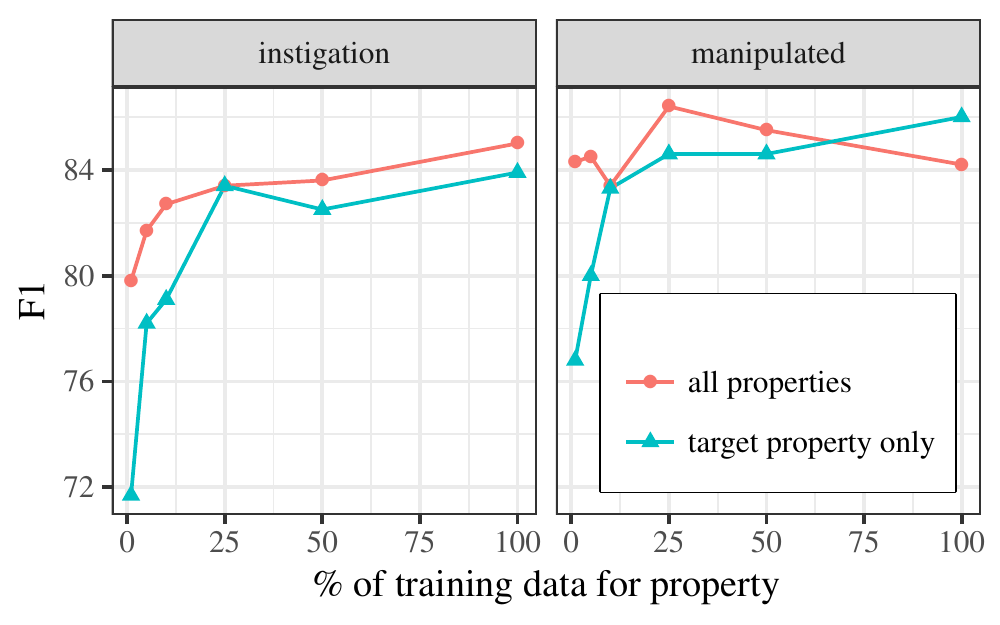}
  \vspace{-5mm}
  \caption{ Effect of using only a fraction of the training data for a property while either ignoring or co-training with the full training data for the other SPR1 properties. Measurements at 1\%, 5\%, 10\%, 25\%, 50\%, and 100\%.}
  \label{fig:ablation}
  \vspace{-5mm}
\end{figure}

\noindent{\bf Learning New SPR
Properties} One motivation for the decompositional approach adopted by
SPRL is the ability to incrementally build up an inventory of
annotated properties according to need and budget.
Here we investigate (1) the degree to which having less training data
for a single property degrades our F1 for that property on held-out
data and (2) the effect on degradation of concurrent training with the
other properties.  We focus on two properties only:
\textsc{instigation}, a canonical example of a proto-agent property,
and \textsc{manipulated}, which is a proto-patient property. For each we consider six training set sizes (1, 5, 10, 25, 50 and 100
percent of the instances). Starting with the same randomly initialized
BiLSTM\footnote{Note that this experiment does not make use of MT
pretraining as was used for Table \ref{tab:spr1_breakdown_test}, to best highlight the impact of parameter sharing across attributes.}, we
consider two training scenarios: (1) ignoring the remaining properties
or (2) including the model's loss on other properties with a weight of
$\lambda = 0.1$ in the training objective.

Results are presented in Figure \ref{fig:ablation}.  We see that, in
every case, most of the performance is achieved with only 25\% of the
training data. The curves also suggest that training simultaneously on
all SPR properties allows the model to learn the target property
more quickly (i.e., with fewer training samples) than if trained
on that property in isolation. For example, at 5\% of the training
training data, the ``all properties'' models are achieving roughly the
same F1 on their respective target property as the ``target property
only'' models achieves at 50\% of the data.\footnote{As we observed
the same trend more clearly on the dev set, we suspect some
over-fitting to the development data which was used for independently
select a stopping epoch for each of the plotted points.}  As the SPR
properties currently annotated are by no means semantically
exhaustive,\footnote{E.g., annotations do not include any questions relating to the \emph{origin} or \emph{destination} of an event.} this experiment indicates that future annotation efforts
may be well served by favoring breadth over depth, collecting smaller
numbers of examples for a larger set of attributes.

\section{Conclusion}
\label{sec:conclusion}
Inspired by: (1) the SPR decomposition of predicate-argument relations into overlapping feature bundles and (2) the neo-Davidsonian formalism for variable-arity predicates, we have proposed
a straightforward extension to a BiLSTM classification framework in
which the states of pre-identified predicate and argument tokens are
pairwise concatenated and used as the target for SPR prediction.  We
have shown that our \emph{Neural-Davidsonian} model outperforms the prior state of the art
in aggregate and showed especially large gains for properties of {\sc
changed-possession}, {\sc stationary}, and {\sc location}. Our
architecture naturally supports discrete or continuous label paradigms, lends itself to multi-task initialization or concurrent training, and allows for
parameter sharing across properties. We demonstrated this
sharing may be useful when some properties are only sparsely annotated
in the training data, which is suggestive of future work in efficiently increasing the range of annotated SPR property types.

\section*{Acknowledgments}
This research was supported by the JHU HLTCOE, DARPA AIDA, and NSF GRFP (Grant No. DGE-1232825).  The U.S. Government is authorized to reproduce and distribute reprints for Governmental purposes. The views and conclusions contained in this publication are those of the authors and should not be interpreted as representing official policies or endorsements of DARPA, NSF, or the U.S. Government.

\bibliography{nds}

\begin{thebibliography}{42}
\expandafter\ifx\csname natexlab\endcsname\relax\def\natexlab#1{#1}\fi

\bibitem[{Abend and Rappoport(2017)}]{abend2017state}
Omri Abend and Ari Rappoport. 2017.
\newblock The state of the art in semantic representation.
\newblock In \emph{Proceedings of the 55th Annual Meeting of the Association
  for Computational Linguistics (Volume 1: Long Papers)}, volume~1, pages
  77--89.

\bibitem[{Bahdanau et~al.(2014)Bahdanau, Cho, and Bengio}]{bahdanau2014neural}
Dzmitry Bahdanau, Kyunghyun Cho, and Yoshua Bengio. 2014.
\newblock Neural machine translation by jointly learning to align and
  translate.
\newblock \emph{arXiv preprint arXiv:1409.0473}.

\bibitem[{Baker et~al.(1998)Baker, Fillmore, and Lowe}]{baker1998berkeley}
Collin~F Baker, Charles~J Fillmore, and John~B Lowe. 1998.
\newblock The berkeley framenet project.
\newblock In \emph{Proceedings of the 36th Annual Meeting of the Association
  for Computational Linguistics and 17th International Conference on
  Computational Linguistics-Volume 1}, pages 86--90. Association for
  Computational Linguistics.

\bibitem[{Bies et~al.(2012)Bies, Mott, Warner, and Kulick}]{bies2012english}
Ann Bies, Justin Mott, Colin Warner, and Seth Kulick. 2012.
\newblock English web treebank.
\newblock \emph{Linguistic Data Consortium, Philadelphia, PA}.

\bibitem[{Bonial et~al.(2014)Bonial, Bonn, Conger, Hwang, and
  Palmer}]{BONIAL14.1012}
Claire Bonial, Julia Bonn, Kathryn Conger, Jena~D. Hwang, and Martha Palmer.
  2014.
\newblock Propbank: Semantics of new predicate types.
\newblock In \emph{Proceedings of the Ninth International Conference on
  Language Resources and Evaluation (LREC'14)}, Reykjavik, Iceland. European
  Language Resources Association (ELRA).

\bibitem[{Callison-Burch et~al.(2009)Callison-Burch, Koehn, Monz, and
  Schroeder}]{callisonburch-EtAl:2009:WMT-09}
Chris Callison-Burch, Philipp Koehn, Christof Monz, and Josh Schroeder. 2009.
\newblock Findings of the 2009 {W}orkshop on {S}tatistical {M}achine
  {T}ranslation.
\newblock In \emph{Proceedings of the Fourth Workshop on Statistical Machine
  Translation}, pages 1--28, Athens, Greece. Association for Computational
  Linguistics.

\bibitem[{Casta{\~n}eda(1967)}]{castaneda-67}
Hector~Neri Casta{\~n}eda. 1967.
\newblock Comment on d. davidson’s "the logical forms of action sentences".
\newblock In N.~Rescher, editor, \emph{The Logic of Decision and Action}.
  University of Pittsburgh Press, Pittsburgh.

\bibitem[{Collobert and Weston(2008)}]{Collobert:2008:UAN:1390156.1390177}
Ronan Collobert and Jason Weston. 2008.
\newblock A unified architecture for natural language processing: Deep neural
  networks with multitask learning.
\newblock In \emph{Proceedings of the 25th International Conference on Machine
  Learning}, ICML '08, pages 160--167, New York, NY, USA. ACM.

\bibitem[{Davidson(1967)}]{davidson-67}
Donald Davidson. 1967.
\newblock The logical forms of action sentences.
\newblock In N.~Rescher, editor, \emph{The Logic of Decision and Action}.
  University of Pittsburgh Press, Pittsburgh.

\bibitem[{Dowty(1991)}]{dowty_thematic_1991}
David Dowty. 1991.
\newblock Thematic proto-roles and argument selection.
\newblock \emph{Language}, 67(3):547--619.

\bibitem[{Fellbaum(1998)}]{fellbaum1998_wordnet}
Christiane Fellbaum. 1998.
\newblock \emph{WordNet: An Electronic Lexical Database}.
\newblock Bradford Books.

\bibitem[{Ganitkevitch et~al.(2013)Ganitkevitch, Van~Durme, and
  Callison-Burch}]{ganitkevitch-vandurme-callisonburch:2013:NAACL-HLT}
Juri Ganitkevitch, Benjamin Van~Durme, and Chris Callison-Burch. 2013.
\newblock Ppdb: The paraphrase database.
\newblock In \emph{Proceedings of the 2013 Conference of the North American
  Chapter of the Association for Computational Linguistics: Human Language
  Technologies}, pages 758--764, Atlanta, Georgia. Association for
  Computational Linguistics.

\bibitem[{Graves et~al.(2013)Graves, Jaitly, and Mohamed}]{graves_hybrid_2013}
Alex Graves, Navdeep Jaitly, and Abdel-rahman Mohamed. 2013.
\newblock Hybrid speech recognition with deep bidirectional {LSTM}.
\newblock In \emph{Automatic {Speech} {Recognition} and {Understanding}
  ({ASRU}), 2013 {IEEE} {Workshop} on}, pages 273--278. IEEE.

\bibitem[{Hashimoto et~al.(2017)Hashimoto, Tsuruoka, Socher
  et~al.}]{hashimoto2017joint}
Kazuma Hashimoto, Yoshimasa Tsuruoka, Richard Socher, et~al. 2017.
\newblock A joint many-task model: Growing a neural network for multiple nlp
  tasks.
\newblock In \emph{Proceedings of the 2017 Conference on Empirical Methods in
  Natural Language Processing}, pages 1923--1933.

\bibitem[{He et~al.(2017)He, Lee, Lewis, and Zettlemoyer}]{he-EtAl:2017:Long3}
Luheng He, Kenton Lee, Mike Lewis, and Luke Zettlemoyer. 2017.
\newblock Deep semantic role labeling: What works and whatâ€™s next.
\newblock In \emph{Proceedings of the 55th Annual Meeting of the Association
  for Computational Linguistics (Volume 1: Long Papers)}, pages 473--483,
  Vancouver, Canada. Association for Computational Linguistics.

\bibitem[{Hochreiter and Schmidhuber(1997)}]{hochreiter_long_1997}
Sepp Hochreiter and Jürgen Schmidhuber. 1997.
\newblock Long short-term memory.
\newblock \emph{Neural computation}, 9(8):1735--1780.

\bibitem[{Howard and Ruder(2018)}]{howard-ruder:2018:Long}
Jeremy Howard and Sebastian Ruder. 2018.
\newblock Universal language model fine-tuning for text classification.
\newblock In \emph{Proceedings of the 56th Annual Meeting of the Association
  for Computational Linguistics (Volume 1: Long Papers)}, pages 328--339,
  Melbourne, Australia. Association for Computational Linguistics.

\bibitem[{Ide and Pustejovsky(2017)}]{ide2017handbook}
N.~Ide and J.~Pustejovsky. 2017.
\newblock \emph{Handbook of Linguistic Annotation}.
\newblock Springer Netherlands.

\bibitem[{Kako(2006)}]{kako2006thematic}
Edward Kako. 2006.
\newblock Thematic role properties of subjects and objects.
\newblock \emph{Cognition}, 101(1):1--42.

\bibitem[{Kingma and Ba(2014)}]{DBLP:journals/corr/KingmaB14}
Diederik~P. Kingma and Jimmy Ba. 2014.
\newblock Adam: {A} method for stochastic optimization.
\newblock \emph{CoRR}, abs/1412.6980.

\bibitem[{Klein et~al.(2017)Klein, Kim, Deng, Senellart, and Rush}]{opennmt}
Guillaume Klein, Yoon Kim, Yuntian Deng, Jean Senellart, and Alexander~M. Rush.
  2017.
\newblock Open{NMT}: Open-source toolkit for neural machine translation.
\newblock In \emph{Proc. ACL}.

\bibitem[{Klerke et~al.(2016)Klerke, Goldberg, and
  S{\o}gaard}]{klerke-goldberg-sogaard:2016:N16-1}
Sigrid Klerke, Yoav Goldberg, and Anders S{\o}gaard. 2016.
\newblock Improving sentence compression by learning to predict gaze.
\newblock In \emph{Proceedings of the 2016 Conference of the North American
  Chapter of the Association for Computational Linguistics: Human Language
  Technologies}, pages 1528--1533, San Diego, California. Association for
  Computational Linguistics.

\bibitem[{Lafferty et~al.(2001)Lafferty, McCallum, and
  Pereira}]{lafferty2001conditional}
John~D. Lafferty, Andrew McCallum, and Fernando C.~N. Pereira. 2001.
\newblock Conditional random fields: Probabilistic models for segmenting and
  labeling sequence data.
\newblock In \emph{Proceedings of the Eighteenth International Conference on
  Machine Learning}, ICML '01, pages 282--289, San Francisco, CA, USA. Morgan
  Kaufmann Publishers Inc.

\bibitem[{Lample et~al.(2016)Lample, Ballesteros, Subramanian, Kawakami, and
  Dyer}]{lample-EtAl:2016:N16-1}
Guillaume Lample, Miguel Ballesteros, Sandeep Subramanian, Kazuya Kawakami, and
  Chris Dyer. 2016.
\newblock Neural architectures for named entity recognition.
\newblock In \emph{Proceedings of the 2016 Conference of the North American
  Chapter of the Association for Computational Linguistics: Human Language
  Technologies}, pages 260--270, San Diego, California. Association for
  Computational Linguistics.

\bibitem[{Luong et~al.(2015)Luong, Pham, and
  Manning}]{luong-pham-manning:2015:EMNLP}
Minh-Thang Luong, Hieu Pham, and Christopher~D. Manning. 2015.
\newblock Effective approaches to attention-based neural machine translation.
\newblock In \emph{Proceedings of the 2015 Conference on Empirical Methods in
  Natural Language Processing}, pages 1412--1421, Lisbon, Portugal. Association
  for Computational Linguistics.

\bibitem[{Luong et~al.(2016)Luong, Le, Sutskever, Vinyals, and
  Kaiser}]{luong2016multiseq}
Thang Luong, Quoc~V. Le, Ilya Sutskever, Oriol Vinyals, and Lukasz Kaiser.
  2016.
\newblock Multi-task sequence to sequence learning.
\newblock In \emph{International Conference on Learning Representations}.

\bibitem[{Ma and Hovy(2016)}]{ma-hovy:2016:P16-1}
Xuezhe Ma and Eduard Hovy. 2016.
\newblock End-to-end sequence labeling via bi-directional lstm-cnns-crf.
\newblock In \emph{Proceedings of the 54th Annual Meeting of the Association
  for Computational Linguistics (Volume 1: Long Papers)}, pages 1064--1074,
  Berlin, Germany. Association for Computational Linguistics.

\bibitem[{Madnani et~al.(2010)Madnani, Boyd-Graber, and Resnik}]{madnani-2010}
Nitin Madnani, Jordan Boyd-Graber, and Philip Resnik. 2010.
\newblock Measuring transitivity using untrained annotators.
\newblock In \emph{Proceedings of the NAACL HLT 2010 Workshop on Creating
  Speech and Language Data with Amazon’s Mechanical Turk}.

\bibitem[{McCann et~al.(2017)McCann, Bradbury, Xiong, and
  Socher}]{NIPS2017_7209}
Bryan McCann, James Bradbury, Caiming Xiong, and Richard Socher. 2017.
\newblock Learned in translation: Contextualized word vectors.
\newblock In I.~Guyon, U.~V. Luxburg, S.~Bengio, H.~Wallach, R.~Fergus,
  S.~Vishwanathan, and R.~Garnett, editors, \emph{Advances in Neural
  Information Processing Systems 30}, pages 6294--6305. Curran Associates, Inc.

\bibitem[{Mou et~al.(2016)Mou, Meng, Yan, Li, Xu, Zhang, and
  Jin}]{mou2016transferable}
Lili Mou, Zhao Meng, Rui Yan, Ge~Li, Yan Xu, Lu~Zhang, and Zhi Jin. 2016.
\newblock How transferable are neural networks in nlp applications?
\newblock In \emph{Proceedings of the 2016 Conference on Empirical Methods in
  Natural Language Processing}, pages 479--489.

\bibitem[{Palmer et~al.(2005)Palmer, Gildea, and
  Kingsbury}]{palmer2005_propbank}
Martha Palmer, Daniel Gildea, and Paul Kingsbury. 2005.
\newblock The proposition bank: An annotated corpus of semantic roles.
\newblock \emph{Computational Linguistics}, 31(1):71--106.

\bibitem[{Parsons(1995)}]{parsons1995thematic}
Terence Parsons. 1995.
\newblock Thematic relations and arguments.
\newblock \emph{Linguistic Inquiry}, pages 635--662.

\bibitem[{Peters et~al.(2018)Peters, Neumann, Iyyer, Gardner, Clark, Lee, and
  Zettlemoyer}]{peters-EtAl:2018:N18-1}
Matthew Peters, Mark Neumann, Mohit Iyyer, Matt Gardner, Christopher Clark,
  Kenton Lee, and Luke Zettlemoyer. 2018.
\newblock Deep contextualized word representations.
\newblock In \emph{Proceedings of the 2018 Conference of the North American
  Chapter of the Association for Computational Linguistics: Human Language
  Technologies, Volume 1 (Long Papers)}, pages 2227--2237, New Orleans,
  Louisiana. Association for Computational Linguistics.

\bibitem[{Poliak et~al.(2018)Poliak, Belinkov, Glass, and
  Van~Durme}]{poliak2018evaluation}
Adam Poliak, Yonatan Belinkov, James Glass, and Benjamin Van~Durme. 2018.
\newblock On the evaluation of semantic phenomena in neural machine translation
  using natural language inference.
\newblock In \emph{Proceedings of the 2018 Conference of the North American
  Chapter of the Association for Computational Linguistics: Human Language
  Technologies, Volume 2 (Short Papers)}, volume~2, pages 513--523.

\bibitem[{Reisinger et~al.(2015)Reisinger, Rudinger, Ferraro, Harman, Rawlins,
  and Van~Durme}]{TACL674}
Drew Reisinger, Rachel Rudinger, Francis Ferraro, Craig Harman, Kyle Rawlins,
  and Benjamin Van~Durme. 2015.
\newblock Semantic proto-roles.
\newblock \emph{Transactions of the Association for Computational Linguistics},
  3:475--488.

\bibitem[{Rudinger et~al.(2018)Rudinger, White, and {Van
  Durme}}]{rudinger-EtAl:2018:N18}
Rachel Rudinger, Aaron~Steven White, and Benjamin {Van Durme}. 2018.
\newblock Neural models of factuality.
\newblock In \emph{Proceedings of the 2018 Conference of the North American
  Chapter of the Association for Computational Linguistics: Human Language
  Technologies}, New Orleans, Louisiana. Association for Computational
  Linguistics.

\bibitem[{Saur{\'i} and Pustejovsky(2009)}]{Sauri2009}
Roser Saur{\'i} and James Pustejovsky. 2009.
\newblock Factbank: a corpus annotated with event factuality.
\newblock \emph{Language Resources and Evaluation}, 43(3):227.

\bibitem[{Schuster and Manning(2016)}]{SCHUSTER16.779}
Sebastian Schuster and Christopher~D. Manning. 2016.
\newblock Enhanced english universal dependencies: An improved representation
  for natural language understanding tasks.
\newblock In \emph{Proceedings of the Tenth International Conference on
  Language Resources and Evaluation (LREC 2016)}, Paris, France. European
  Language Resources Association (ELRA).

\bibitem[{Silveira et~al.(2014)Silveira, Dozat, de~Marneffe, Bowman, Connor,
  Bauer, and Manning}]{silveira14gold}
Natalia Silveira, Timothy Dozat, Marie-Catherine de~Marneffe, Samuel Bowman,
  Miriam Connor, John Bauer, and Christopher~D. Manning. 2014.
\newblock A gold standard dependency corpus for {E}nglish.
\newblock In \emph{Proceedings of the Ninth International Conference on
  Language Resources and Evaluation (LREC-2014)}.

\bibitem[{Teichert et~al.(2017)Teichert, Poliak, Van~Durme, and
  Gormley}]{teichert2017sprl}
Adam Teichert, Adam Poliak, Benjamin Van~Durme, and Matthew~R Gormley. 2017.
\newblock Semantic proto-role labeling.
\newblock In \emph{Thirty-First AAAI Conference on Artificial Intelligence
  (AAAI-17)}.

\bibitem[{White et~al.(2016)White, Reisinger, Sakaguchi, Vieira, Zhang,
  Rudinger, Rawlins, and Van~Durme}]{uds2016}
Aaron~Steven White, Drew Reisinger, Keisuke Sakaguchi, Tim Vieira, Sheng Zhang,
  Rachel Rudinger, Kyle Rawlins, and Benjamin Van~Durme. 2016.
\newblock Universal decompositional semantics on universal dependencies.
\newblock In \emph{Proceedings of the 2016 Conference on Empirical Methods in
  Natural Language Processing}, pages 1713--1723, Austin, Texas. Association
  for Computational Linguistics.

\bibitem[{Zhang et~al.(2017)Zhang, Rudinger, and {Van
  Durme}}]{zhang-EtAl:2017:IWCS}
Sheng Zhang, Rachel Rudinger, and Benjamin {Van Durme}. 2017.
\newblock {An Evaluation of PredPatt and Open IE via Stage 1 Semantic Role
  Labeling}.
\newblock In \emph{Proceedings of the 12th International Conference on
  Computational Semantics (IWCS)}.

\end{thebibliography}
\bibliographystyle{acl_natbib_nourl}

\appendix
\clearpage

\section{Mult-Task Investigation}
\label{sec:multi}
\begin{table}[t!]
\begin{center}
\small
\begin{tabular}{|l|c|l|}
\hline \bf Name & \# & \bf Description \\ \hline
\Reisinger & & Logistic Regr. model, \\
 & & \hspace{.5em} \newcite{TACL674} \\
\AAAI & & CRF model,\\
 & & \hspace{.5em} \newcite{teichert2017sprl} \\
\hline
\SprOneB & 0 & SPR1 basic model \\
\SprOneRand & 0 & \SprOneB, random word embeddings \\
\mtSprOne & 1a & \SprOneB~after MT pretraining \\
\pbSprOne & 1a & \SprOneB~after PB pretraining \\
\mtpbSprOne & 1a & \SprOneB~after MT+PB pretraining\\
\SprOneBSprTwo & 1b & SPR1 and SPR2 concurrently \\
\SprOneBSupersense & 1b & SPR1 and WSD concurrently \\
\mtSprOneBSprTwo & 1b & \SprOneBSprTwo~after MT pretraining\\
\mtSprOneBSupersense & 1b & \SprOneBSupersense~after  MT pretraining\\
\mtSprOneS & 1c & SPR1 scalar after MT pretraining\\
\pbSprOneS & 1c & SPR1 scalar after PB pretraining\\
\psopt & 1d & SPR1 propty-specific model sel.\\
\SprTwo & 3 & SPR2 basic scalar model \\
\mtSprTwo & 3 & \SprTwo~after MT pretraining\\
\pbSprTwo & 3 & \SprTwo~after PB pretraining\\
\mtpbSprTwo & 3 & \SprTwo~after MT+PB pretraining\\
\hline
\end{tabular}
\end{center}
\caption{\label{tab:names} Name and short description of each experimental condition reported. {\sc mt:} indicates pretraining with machine translation; {\sc pb:} indicates pretraining with PropBank SRL.}
\end{table}

Multi-task learning has been found to improve performance on many NLP
tasks, particularly for neural models, and is rapidly becoming
\textit{de rigueur} in the field.  The strategy involves optimizing
for multiple training objectives corresponding to different (but
usually related) tasks.  \newcite{Collobert:2008:UAN:1390156.1390177}
use multi-task learning to train a convolutional neural network to
perform multiple core NLP tasks (POS tagging, named entity
recognition, etc.). Multi-task learning has also been used to improve
sentence compression \cite{klerke-goldberg-sogaard:2016:N16-1},
chunking and dependency parsing \cite{hashimoto2017joint}. Related
work on UDS \cite{uds2016} shows improvements on event factuality prediction with
multi-task learning on BiLSTM models \cite{rudinger-EtAl:2018:N18}.  To
complete the basic experiments reported in the main text, here we
include an investigation of the impact of multi-task learning for
SPRL.

We borrow insights from \newcite{mou2016transferable} who explore
different multi-task strategies for NLP including approach of
initializing a network by training it on a related task (``INIT'')
versus interspersing tasks during training (``MULT''). Here we employ
both of these strategies, referring to them as \textit{pretraining}
and \textit{concurrent} training.  We also use the terminology
\textit{target task} and \textit{auxiliary task} to differentiate the
primary task(s) we are interested in from those that play only a
supporting role in training. In order to tune the impact of auxiliary
tasks on the learned representation, \newcite{luong2016multiseq} use a
{\em mixing parameter}, $\alpha_i$, for each task $i$.  Each parameter
update consists of selecting a task with probability proportional to
its $\alpha_i$ and then performing one update with respect to that
task alone.  They show that the choice of $\alpha$ has a large impact
on the effect of multi-task training, which influences our experiments
here.

Please refer to Appendix \ref{sec:data} for details on the datasets
used in this section.  In particular, with a few exceptions,
\newcite{uds2016} annotates for the same set of properties as
\newcite{TACL674}, but with slightly different protocol and on a
different genre.  However, in this section we treat the two datasets
as if they were separate tasks.  To avoid cluttering the results in the
main text, we exclusively present results there on what we call {\em
  SPR1} which consists of the data from \newcite{TACL674} and the
train/dev/test splits of \newcite{teichert2017sprl}. We refer to the
analogous tasks built on the data and splits of \newcite{uds2016}
using the term SPR2. (We are not aware of any prior published results
on property prediction for the SPR2.)

In addition to the binary and scalar SPR architectures outlined in
Section \ref{sec:model} of the main paper, we also considered
concurrently training the BiLSTM on a fine-grained word-sense
disambiguation task or on joint SPR1 and SPR2 prediction.  We also
experimented with using machine translation and PropBank SRL to
initialize the parameters of the BiLSTM.  Preliminary experimentation
on dev data with other combinations helped prune down the set of
interesting experiments to those listed in Table \ref{tab:names} which
assigns names to the models explored here.  Our ablation study in
Section \ref{sec:experiments} of the main paper uses the model named
$\SprOneB$ while the other results in the main paper correspond to
$\mtSprOne$ in the case of binary prediction and $\mtSprOneS$ in the
case of scalar prediction.  After detailing the additional components
used for pretraining or concurrent training, we  present aggregate
results and for the best performing models (according to dev) we 
present property-level aggregate
results.

\subsection{Auxiliary Tasks}
Each auxiliary task is implemented in the form of a task-specific decoder with access to the hidden states computed by the shared BiLSTM encoder. In this way, the losses from these tasks backpropagate through the BiLSTM. Here we describe each task-specific decoder.

\paragraph{PropBank Decoder} The network architecture for the
auxiliary task of predicting abstract role types in PropBank is nearly
identical to the architecture for SPRL described in Section
\ref{sec:model} of the main paper. The main difference is that the
PropBank task is a single-label, categorical classification task.

\begin{center}
$\text{P}(\text{role}_i|\bm{h}_{ea})=\text{softmax}_i\left(\bm{W}_{\text{propbank}}\left[\bm{h}_{ea}\right]\right)$
\end{center}

The loss from this decoder is the negative log of the probability
assigned to the correct label.

\paragraph{Supersense Decoder}
The word sense disambiguation decoder computes a probability
distribution over 26 WordNet supersenses with a simple single-layer
feedforward network:
\begin{align*}
\text{P}(\text{supersense}_i|\bm{h}_a) &= \text{softmax}_i(\bm{W}\left[\bm{h_a}\right])
\end{align*}

\noindent where $\bm{W} \in \mathds{R}^{1200\times 26}$ and $\bm{h}_a$ is the RNN hidden state corresponding to the argument head token we wish to disambiguate.
Since the gold label in the supersense prediction task is a \textit{distribution} over supersenses, the loss from this decoder is the cross-entropy between its predicted distribution and the gold distribution. 

\paragraph{French Translation Decoder}

Given the encoder hidden states, the goal of translation is to
generate the reference sequence of tokens $Y=y_1,\cdots,y_n$ in the
target language, i.e., French.  We employ the standard decoder
architecture for neural machine translation.  At each time step $i$,
the probability distribution of the decoded token $y_i$ is defined as:
\begin{equation*}
    P(y_i) = \text{softmax}\big(\text{tanh}(\bm{W_\textrm{fr}}\big[\bm{s_i};\bm{c_i}\big] + \bm{b_\textrm{fr}})\big)
\end{equation*}
where $\bm{W_\textrm{fr}}$ is a transform matrix, and
$\bm{b_\textrm{fr}}$ is a bias. The inputs are the decoder hidden
state $\bm{s_i}$ and the context vector $\bm{c_i}$.  The decoder
hidden state \bm{$s_i$} is computed by:
\begin{equation*}
    \bm{s_i} = \textsc{rnn}(\bm{y_{i-1}}, \bm{s_{i-1}}) \label{eq:decoder-rnn}
\end{equation*}
where \textsc{rnn} is a recurrent neural network using $L$-layer stacked LSTM,
$\bm{y_{i-1}}$ is the word embedding of token $y_{i-1}$, and $\bm{s_0}$ is initialized by
the last encoder left-to-right hidden state.

The context vector $\bm{c_i}$ is computed by an attention mechanism
\cite{bahdanau2014neural,luong-pham-manning:2015:EMNLP},
\begin{align*}
    \bm{c_{i}} & = \sum_{t}\alpha_{i,t}\bm{h_t}, \\
    \alpha_{i,t} & = \frac{\exp{\big(\bm{s_{i}}^\top (\bm{W_{\alpha}h_t} + \bm{b_\alpha})\big))}}{\sum_{k}\exp{\big(\bm{s_{i}}^\top (\bm{W_{\alpha}h_{k}} + \bm{b_\alpha})\big)}},
\end{align*}
where $\bm{W_\alpha}$ is a transform matrix and $\bm{b_\alpha}$ is a
bias. The loss is the negative log-probability of the decoded
sequence.

\subsection{Results}

In this section, we present a series of experiments using different
components of the neural architecture described in Section \ref{sec:model},
with various training regimes.  Each experimental setting is given a
name (in \textsc{smallcaps}) and summarized in Table \ref{tab:names}.
Unless otherwise stated, the target task is SPR1 (classification). To
ease comparison, we include results from the main paper as well as
additional results.

\begin{table}
\begin{center}
\begin{tabular}{lcc}
\toprule
{} &  micro-F1 &  macro-F1 \\
\midrule
\Reisinger  &      71.0 &  55.4$^\star$ \\
\AAAI       &      81.7 &  65.9$^\star$ \\
\hline
\SprOneRand    &      77.7 &   57.3 \\
\SprOneB    &      \bf{82.2} &   69.3 \\
\mtSprOneB    &      \bf{83.3} &   71.1 \\
\pbSprOneB    &     \bf{82.3} &   67.9 \\
\mtpbSprOneB  &     \bf{82.8} &   70.9 \\
\SprOneBSprTwo    &    \bf{83.3} &   70.4 \\
\SprOneBSupersense    &       \bf{81.9} &   67.9 \\
\mtSprOneBSprTwo    &      \bf{83.2} &   70.0 \\
\mtSprOneBSupersense    &     \bf{81.8} &   67.4 \\
\psopt    &     \bf{82.9} &   69.5 \\
\bottomrule
\end{tabular}
\end{center}
\caption{\label{tab:spr1c_overall_test} Overall test performance for all settings described in Experiments 1 and 1a-d. The target task is SPR1 as binary classification. Micro- and macro-F1 are computed over all properties. ($^\star$Baseline macro-F1 scores are computed from property-specific precision and recall values in \newcite{teichert2017sprl} and may introduce rounding errors.)}
\end{table}

\begin{table}
\begin{center}
\small
\begin{tabular}{lllcc}
\toprule
{} &  \!\AAAI\! & \!\SprOneB\! & \!\mtSprOneB & \!\SprOneBSprTwo\! \\
\midrule
instigation                       &  85.6& 84.6 &     \bf{88.6} &       85.6 \\
\rowcolor{\rowc}volition          &  86.4& \bf{87.9} &     \bf{88.1} &       \bf{88.0} \\
awareness                         &  87.3& \bf{88.3} &     \bf{89.9} &       \bf{88.4} \\
\rowcolor{\rowc}sentient          &  85.6& \bf{89.6} &     \bf{90.6} &       \bf{90.0} \\
physically existed                &  76.4& \bf{82.3} &     \bf{82.7} &       \bf{80.2} \\
\rowcolor{\rowc}existed before    &  84.8& \bf{86.0} &     \bf{85.1} &       \bf{86.8} \\
existed during                    &  95.1& 94.2 &     95.0 &       94.8 \\
\rowcolor{\rowc}existed after     &  87.5& 86.9 &     85.9 &       87.5 \\
created                           &  44.4& \bf{46.6} &     39.7 &       \bf{51.6} \\
\rowcolor{\rowc}destroyed         &  0.0&  \bf{11.1} &     \bf{24.2} &        \bf{6.1} \\
changed                           &  67.8& 67.4 &     \bf{70.7} &       \bf{68.1} \\
\rowcolor{\rowc}changed state     &  66.1& \bf{66.8} &     \bf{71.0} &       \bf{67.1} \\
changed possession                &  38.8& \bf{57.1} &     \bf{58.0} &       \bf{63.7} \\
\rowcolor{\rowc}changed location  &  35.6& \bf{60.0} &     \bf{45.7} &       \bf{52.9} \\
stationary                        &  21.4& \bf{43.2} &     \bf{47.4} &       \bf{53.1} \\
\rowcolor{\rowc}location          &  18.5& \bf{46.9} &     \bf{53.8} &       \bf{53.6} \\
physical contact                  &  40.7& \bf{52.7} &     \bf{47.2} &       \bf{54.7} \\
\rowcolor{\rowc}manipulated       &  86.0& 82.2 &     \bf{86.8} &       \bf{86.7} \\
\hline
micro f1                          &  81.7 & \bf{82.2} &     \bf{83.3} &       \bf{83.3} \\
\rowcolor{\rowc}macro f1          &  65.9 & \bf{69.3} &     \bf{71.1} &    \bf{70.4} \\
\bottomrule           
\end{tabular}
\end{center}
\caption{\label{tab:spr1c_breakdown_test} Breakdown by property of binary classification F1 on SPR1. All new results outperforming prior work (\AAAI) in bold.}
\end{table}

\paragraph{Experiment 0: Embeddings} By default, all models reported in this paper employ pretrained word embeddings (GloVe). In this experiment we replaced the pretrained embeddings in the vanilla SPR1 model (\SprOneB) with randomly initialized word embeddings (\SprOneRand). The results (Table \ref{tab:spr1c_overall_test}) reveal substantial gains from the use of pretrained embeddings; this is likely due to the comparatively small size of the SPR1 training data.

\paragraph{Experiment 1a: Multi-task Pretraining} We pretrained the
BiLSTM encoder with two separate auxiliary tasks: \textbf{French
  Translation} and \textbf{PropBank Role Labeling}.  There are three
settings: (1) Translation pretraining only (\mtSprOne), (2) PropBank
pretraining only (\pbSprOne), and (3) Translation pretraining followed
by PropBank pretraining (\mtpbSprOne).  In each case, after
pretraining, the SPRL decoder is trained end-to-end, as in Experiment
0 (on SPR1 data).

\paragraph{Experiment 1b: Multi-task Concurrent} One auxiliary task
(\textbf{Supersense} or \textbf{SPR2}) is trained concurrently with
SPR1 training. In one epoch of training, a training example is sampled
at random (without replacement) from either task until all training
instances have been sampled.  The loss from the auxiliary task (which,
in both cases, has more training instances than the target SPRL task)
is down-weighted in proportion to ratio of the dataset sizes:
\begin{align*}
\alpha = \frac{|\text{target task}|}{|\text{auxiliary task}|}
\end{align*}
The auxiliary task loss is further down-weighted by a hyperparameter $\lambda\in\{1,10^{-1},10^{-2},10^{-3},10^{-4}\}$ which is chosen based on dev results.
We apply this training regime with the auxiliary task of Supersense prediction (\SprOneSupersense) and the scalar SPR2 prediction task (\SprOneSprTwo), described in Experiment 2.

\paragraph{Experiment 1c: Multi-task Combination}
This setting is identical to Experiment 1b, but includes MT
pretraining (the best-performing pretraining setting on dev), as
described in 1a. Accordingly, the two experiments are
\mtSprOneSupersense\hspace{.1em} and \mtSprOneSprTwo.

\paragraph{Experiment 1d: Property-Specific Model Selection} (\textsc{ps-ms})
Experiments 1a--1c consider a variety of pretraining tasks,
co-training tasks, and weight values, $\lambda$, in an effort to
improve aggregate F1 for SPR1.  However, the SPR properties are
diverse, and we expect to find gains by choosing training settings on
a property-specific basis.  Here, for each property, we select from
the set of models considered in experiments 1a--1c the one that
achieves the highest dev F1 for the target property. We report the
results of applying those property-specific models to the test data.

\begin{table}
\setlength{\tabcolsep}{5pt}
\begin{center}
\small
\begin{tabular}{lccc}
\toprule
                SPR property &      \SprOneS &      \mtSprOneS &      \SprTwo \\
\midrule
                 instigation &         0.835 &           0.858 &        0.590 \\
\rowcolor{\rowc}    volition &         0.869 &           0.882 &        0.837 \\
                   awareness &         0.873 &           0.897 &        0.879 \\
\rowcolor{\rowc}    sentient &         0.917 &           0.925 &         0.880 \\
          physically existed &         0.820 &           0.834 &            - \\
\rowcolor{\rowc} existed before &         0.696 &           0.710 &        0.616 \\
              existed during &         0.666 &           0.673 &        0.358 \\
\rowcolor{\rowc} existed after &         0.612 &           0.619 &        0.478 \\
                     created &         0.540 &           0.549 &            - \\
\rowcolor{\rowc}   destroyed &         0.268 &           0.346 &            - \\
                     changed &         0.619 &           0.592 &            - \\
\rowcolor{\rowc} changed state &         0.616 &           0.604 &        0.352 \\
        changed possession &         0.652 &           0.640 &        0.488 \\
\rowcolor{\rowc} change of location &         0.778 &           0.777 &        0.492 \\
  changed state continuous &             - &               - &        0.373 \\
\rowcolor{\rowc} was for benefit &             - &               - &        0.578 \\
                  stationary &         0.705 &           0.711 &            - \\
\rowcolor{\rowc}    location &         0.627 &           0.619 &            - \\
            physical contact &         0.731 &           0.741 &            - \\
\rowcolor{\rowc} manipulated &         0.715 &           0.737 &            - \\
                    was used &             - &               - &        0.203 \\
\rowcolor{\rowc}   partitive &             - &               - &        0.359 \\
\hline
           macro-avg pearson &         0.743 &           0.753 &        0.591 \\
\bottomrule
\end{tabular}
\end{center}
\caption{\label{tab:scalar_test} SPR1 and SPR2 as scalar prediction tasks. Pearson correlation between predicted and gold values. An earlier version of this paper contained errors in this table that have been corrected in this version; see \url{https://github.com/decomp-sem/neural-sprl}.}
\end{table}
\begin{table}[ht!]
\begin{center}
\begin{tabular}{lrlr}
\toprule
\SprOneS &  0.697 &      \SprTwo &  \bf{0.534} \\
\mtSprOneS    &  \bf{0.706} &    \mtSprTwo &  0.521 \\
\pbSprOneS   &  0.685 &     \pbSprTwo &  0.511 \\
\mtpbSprOneS &  0.675 &  \mtpbSprTwo &  0.508 \\
\bottomrule
\end{tabular}
\end{center}
\caption{\label{tab:spr1s_overall_test} SPR1 and SPR2 as scalar prediction tasks. The overall performance for each experimental setting is reported as the average Pearson correlation over all properties. Highest SPR1 and SPR2 results are in bold. An earlier version of this paper contained errors in this table that have been corrected in this version; see \url{https://github.com/decomp-sem/neural-sprl}.}
\end{table}

\paragraph{Experiment 2: SPR as a scalar task}
In Experiment 2, we trained the SPR decoder to predict properties as scalar instead of binary values.
Performance is measured by Pearson correlation and reported in Tables \ref{tab:spr1s_overall_test} and \ref{tab:scalar_test}.
In this case, we treat SPR1 and SPR2 both as target tasks (separately). By including SPR1 as a target task, we are able to compare (1) SPR as a binary task and a scalar task, as well as (2) SPR1 and SPR2 as scalar tasks.
These results constitute the first reported numbers on SPR2.

We observe a few trends. First, it is generally the case that properties with high F1 on the SPR1 binary task also have high Pearson correlation on the SPR1 scalar task. The higher scoring properties in SPR1 scalar are also generally the higher scoring properties in SPR2 (where the SPR1 and SPR2 properties overlap), with a few notable exceptions, like \textsc{instigation}. Overall, correlation values are lower in SPR2 than SPR1. This may be the case for a few reasons. (1) The underlying data in SPR1 and SPR2 are quite different. The former consists of sentences from the Wall Street Journal via PropBank \cite{palmer2005_propbank}, while the latter consists of sentences from the English Web Treebank \cite{bies2012english} via the Universal Dependencies; (2) certain filters were applied in the construction of the SPR1 dataset to remove instances where, e.g., predicates were embedded in a clause, possibly resulting in an easier task; (3) SPR1 labels came from a single annotator (after determining in pilot studies that annotations from this annotator correlated well with other annotators), where SPR2 labels came from 24 different annotators with scalar labels averaged over two-way redundancy.

\paragraph{Discussion} With SPR1 binary classification as the target task, we see overall
improvements from various multi-task training regimes (Experiments
1a-d, Tables \ref{tab:spr1c_overall_test} and
\ref{tab:spr1c_breakdown_test}), using four different auxiliary tasks:
machine translation into French, PropBank abstract role prediction,
word sense disambiguation (WordNet supersenses), and
SPR2.\footnote{Note that in some cases we treat SPR2 as an auxiliary
  task, and in others, the target task.} These auxiliary tasks exhibit
a loose trade-off in terms of the quantity of available data and the
semantic relatedness of the task: MT is the least related task with
the most available (parallel) data, while SPR2 is the most related
task with the smallest quantity of data. While we hypothesized that
the relatedness of PropBank role labeling and word sense
disambiguation tasks might lead to gains in SPR performance, we did
not see substantial gains in our experiments (\pbSprOneB,
\SprOneBSupersense). We did, however, see improvements over the
target-task only model (\SprOneB) in the cases where we added MT
pretraining (\mtSprOneB) or SPR2 concurrent training
(\SprOneBSprTwo). Interestingly, combining MT pretraining with SPR2
concurrent training yielded no further gains (\mtSprOneBSprTwo).

\section{Data}
\label{sec:data}
\paragraph{SPR1} The SPR1.0 (``SPR1'') dataset introduced by
\newcite{TACL674} contains proto-role annotations on 4,912 Wall Street
Journal sentences from PropBank \cite{palmer2005_propbank}
corresponding to 9,738 predicate-argument pairs with 18 properties
each, in total 175,284 property annotations.  All annotations were
performed by a single, trusted annotator.  Each annotation is a rating
from 1 to 5 indicating the likelihood that the property applies, with
an additional ``N/A'' option if the question of whether the property
holds is nonsensical in the context.

To compare with prior work \cite{teichert2017sprl}, we treat the SPR1
data as a binary prediction task: the values 4 and 5 are mapped to
\textbf{True} (property holds), while the values 1, 2, 3, and ``N/A''
are mapped to \textbf{False} (property does not hold).  In additional
experiments, we move to treating SPR1 as a scalar prediction task; in
this case, ``N/A'' is mapped to 1, and all other annotation values
remain unchanged.

\noindent\textbf{SPR2} The second SPR release \cite{uds2016} contains annotations on 2,758 sentences from the English Web Treebank (EWT) \cite{bies2012english} portion of the Universal Dependencies (v1.2) \cite{silveira14gold}\footnote{We exclude the SPR2 pilot data; if included, the SPR2 release contains annotations for 2,793 sentences.}, corresponding to 6,091 predicate-argument pairs. With 14 proto-role properties each, there are a total of 85,274 annotations, with two-way redundancy. As in SPR1, the value of each annotation is an integral value 1-5 or ``N/A.''
We treat SPR2 as a scalar prediction task, first mapping ``N/A'' to 1,
and then averaging the two-way redundant annotation values to a single
value.

\paragraph{Word Sense Disambiguation} Aligned with proto-role property
annotations in the SPR2 release are word sense disambiguation
judgments for the head tokens of arguments. Candidate word senses
(fine-grained) from WordNet \cite{fellbaum1998_wordnet} were presented
to Mechanical Turk workers (at least three annotators per instance),
who selected every applicable sense of the word in the given
context. In this work, we map the fine-grained word senses to one of
26 coarse-grained WordNet noun supersenses (e.g.,
\texttt{noun.animal}, \texttt{noun.event}, \texttt{noun.quantity},
etc.). In many cases, a word may be mapped to more than one
supersense. We treat the supersense label on a word as a distribution
over supersenses, where the probability assigned to one supersense is
proportional to the number of annotators that (indirectly) selected
that supersense. In practice, the entropy of these resulting
supersense distributions is low, with an average perplexity of 1.42.

\paragraph{PropBank} The PropBank project consists of
predicate-argument annotations over corpora for which gold Penn
TreeBank-style constituency parses are available.  We use the Unified
PropBank release \cite{BONIAL14.1012, ide2017handbook}, which contains
annotations over OntoNotes as well as the English Web TreeBank
(EWT). Each predicate in each corpus is annotated for word sense, and
each argument of each predicate is given a label such as ARG0, ARG1,
etc., where the interpretation of the label is defined relative to the
word sense.  We use PropBank Frames to map these sense-specific labels
to 16 sense-independent labels such as PAG (proto-agent), PPT
(proto-patient), etc., and then formulate a task to predict the
abstracted labels. Because our model requires knowledge of predicate
and argument head words, we ran the Stanford Universal Dependencies
converter \cite{SCHUSTER16.779} over the gold constituency parses to
obtain Universal Dependency parses, which were then processed by the
PredPatt framework \cite{zhang-EtAl:2017:IWCS,uds2016} to identify
head words.

\paragraph{English-French Data} The $10^9$ French-English parallel
corpus \cite{callisonburch-EtAl:2009:WMT-09} contains 22,520,376
French-English sentence pairs, made up of 811,203,407 French words and
668,412,817 English words.  The corpus was constructed by crawling the
websites of international organizations such as the Canadian
government, the European Union, and the United Nations.

\end{document}